\documentclass[sigconf, anonymous=false]{acmart}

\usepackage{bm}

\usepackage{multirow}
\usepackage{subfig}
\usepackage{algorithm}
\usepackage{algorithmic}

\AtBeginDocument{%
  \providecommand\BibTeX{{%
    \normalfont B\kern-0.5em{\scshape i\kern-0.25em b}\kern-0.8em\TeX}}}

\setcopyright{rightsretained}
\copyrightyear{2021}
\acmYear{2021}
\setcopyright{rightsretained}
\acmConference[KDD '21]{Proceedings of the 27th ACM SIGKDD Conference on Knowledge Discovery and Data Mining}{August 14--18, 2021}{Virtual Event, Singapore}
\acmBooktitle{Proceedings of the 27th ACM SIGKDD Conference on Knowledge Discovery and Data Mining (KDD '21), August 14--18, 2021, Virtual Event, Singapore}\acmDOI{10.1145/3447548.3467382}
\acmISBN{978-1-4503-8332-5/21/08}

\acmSubmissionID{rst2800}
\begin{document}
\fancyhead{}
\title{Uncertainty-Aware Reliable Text Classification}

\author{Yibo Hu}

\affiliation{%
  \institution{The University of Texas at Dallas}
  \city{Richardson}
  \state{TX}
  \country{USA}
  \postcode{75080}
}
\email{yibo.hu@utdallas.edu}

\author{Latifur Khan}

\affiliation{%
  \institution{The University of Texas at Dallas}
  \city{Richardson}
  \state{TX}
  \country{USA}
  \postcode{75080}
}
\email{lkhan@utdallas.edu}

\begin{abstract}

Deep neural networks have significantly contributed to the success in predictive accuracy for classification tasks. However, they tend to make over-confident predictions in real-world settings, where domain shifting and out-of-distribution (OOD) examples exist. Most research on uncertainty estimation focuses on computer vision because it provides visual validation on uncertainty quality. However, few have been presented in the natural language process domain. Unlike Bayesian methods that indirectly infer uncertainty through weight uncertainties, current evidential uncertainty-based methods explicitly model the uncertainty of class probabilities through subjective opinions. They further consider inherent uncertainty in data with different root causes, vacuity (i.e., uncertainty due to a lack of evidence) and dissonance (i.e., uncertainty due to conflicting evidence).  In our paper, we firstly apply evidential uncertainty in OOD detection for text classification tasks. We propose an inexpensive framework that adopts both auxiliary outliers and pseudo off-manifold samples to train the model with prior knowledge of a certain class, which has high vacuity for OOD samples. Extensive empirical experiments demonstrate that our model based on evidential uncertainty outperforms other counterparts for detecting OOD examples. Our approach can be easily deployed to traditional recurrent neural networks and fine-tuned pre-trained transformers.

\end{abstract}



\maketitle

\section{Introduction} \label{sec:intro}

Deep neural networks have significantly contributed to the success of predictive accuracy for classification tasks in multiple domains. 
However, many applications require confidence in reliability. In real-world settings that contain out-of-distribution (OOD) samples, the model should know when it can not make a confident judgment rather than making an incorrect one. 
Studies show that traditional neural networks easily lead to over-confidence, i.e., a high-class probability in an incorrect class prediction~\cite{guo2017calibration,hein2019relu,ovadia2019canshift}.   
Therefore, calibrated predictive uncertainty is crucial to avoid those risks. 

In this paper, we are interested in qualifying uncertainty to solve OOD detection in text classification as it contains a wide range of Natural Language Processing (NLP) applications~\cite{Chang2020TamingPT,Li2018LearningAN}.
Although fine-tuning pre-trained transformers~\cite{devlin2018bert}
have achieved state-of-the-art accuracy on text classification tasks, they still suffer from the same over-confidence problem of traditional neural networks, making the prediction untrustful~\cite{hendrycks2020pretrained}. 
One partial explanation is over-parameterization~\cite{guo2017calibration}. 
Although transformers are pre-trained on a large corpus and get rich semantic information, it leads to over-confidence easily given limited labeled data during the fine-tuning stage~\cite{kong2020calibrated}. 
Overall, compared to the Computer Vision (CV) domain, there is less work in qualifying uncertainty in the NLP domain.  Among them, there are Bayesian and non-Bayesian methods.

Bayesian models qualify the model uncertainty by Bayesian neural networks (BNNs) \cite{blundell2015weight,louizos2017multiplicative}. 
BNNs explicitly qualify model uncertainty by considering model parameters as distributions. 
Specifically,  BNNs consider probabilistic uncertainty, i.e., aleatoric uncertainty and epistemic uncertainty~\cite{kendall2017uncertainties}. Aleatoric only considers data uncertainty caused by statistical randomness. At the same time, epistemic refers to model uncertainty introduced by limited knowledge or ignorance in collected data. 
Monte Carlo Dropout~\cite{gal2016dropout} is a crucial technique to approximate variational Bayesian inference. It trains and evaluates a neural network with dropout~\cite{srivastava2014dropout} before each layer. 
BNNs have been explored for classification prediction or regression in CV applications. However, there has been less study in the NLP domain.  
Few work~\cite{xiao2019quantifying,van2020predictive,ovadia2019canshift} empirically evaluate uncertainty estimation in text classification.
Other attempts adopt MC Dropout in deep active learning~\cite{shen2017deep,Siddhant2018deep}, sentiment analysis~\cite{Andersen2020wordlevel}, or machine translation~\cite{Zhou2020uncertainty}.

Non-Bayesian approaches use entropy~\cite{shannon1948mathematical} or softmax scores as a measure of uncertainty, which only considers aleatoric uncertainty~\cite{kendall2017uncertainties}. 
OOD detection in text classification using GRU~\cite{chung2014empirical} or LSTM~\cite{schmidhuber1997long} has been studied in \cite{hendrycks2016baseline, hendrycks2018deep}.
\citet{hendrycks2020pretrained} empirically study pre-trained transformers' performance on OOD detection. They point out transformers cannot clearly separate in-distribution (ID) and OOD examples.  
In addition,  OOD detection has also been studied in dialogue systems~\cite{zheng2020out} and document classification~\cite{Zhang2019mitigating,he2020towards}. 
Another line of non-Bayesian methods involves the calibration of probabilities.
Temperature scaling~\cite{guo2017calibration} calibrates softmax probabilities by adding a scalar parameter to each class in a post-processing step.  
\citet{Thulasidasan2019mixup} explore the improvement of calibration and predictive uncertainty of models trained with mix-up \cite{zhang2017mixup} in the NLP domain.  
\citet{kong2020calibrated} use pseudo samples on and off the data manifold for calibration. 

\begin{table*}[htbp]
\centering \small
\caption{Predictive uncertainty of sentiment analysis of restaurant reviews. The model without calibration demonstrates over-confidence. A well-calibrated classifier outputs the same expected probabilities for Case 2 and 3 that have different evidence.}
\label{example}
\begin{tabular}{c|l|l|l|l} 
\toprule
\multicolumn{1}{l|}{Calibrated?} & Test Sentence                                                                                                                  & Probability         & Dirichlet            & Uncertainty 
\\
\hline
No                               & 3. 'Deep learning is data hungry.'                                                                                             & $p$~= [0.99, 0.1]    & doesn't apply & Over-confidence                                                                     \\ 
\hline
\multirow{3}{*}{Yes}             & \begin{tabular}[c]{@{}l@{}}1. 'This was the worst restaurant I have ever had the misfortune of eating at.' \end{tabular}     & $p$ = [0.01, 0.99]  & $\alpha$ = [1,99]    & Low uncertainty                                                   \\ 
\cline{2-5}
                                 & \begin{tabular}[c]{@{}l@{}}2. 'This restaurant is bad. Yet its food is  acceptable considering the low price.' \end{tabular} & $p$ = [0.5, 0.5]    & $\alpha$ = [50, 50]  & \begin{tabular}[c]{@{}l@{}}Conflicting evidence \end{tabular}  \\ 
\cline{2-5}
                                 & 3.~'Deep learning is data hungry.'                                                                                             & $p$ = [0.5, 0.5]    & $\alpha$ = [1, 1]    & \begin{tabular}[c]{@{}l@{}}Lack of evidence \end{tabular}          \\
\bottomrule
\end{tabular}
\end{table*}

\begin{figure*}[htbp]
\centering
\includegraphics[width=0.95\textwidth]{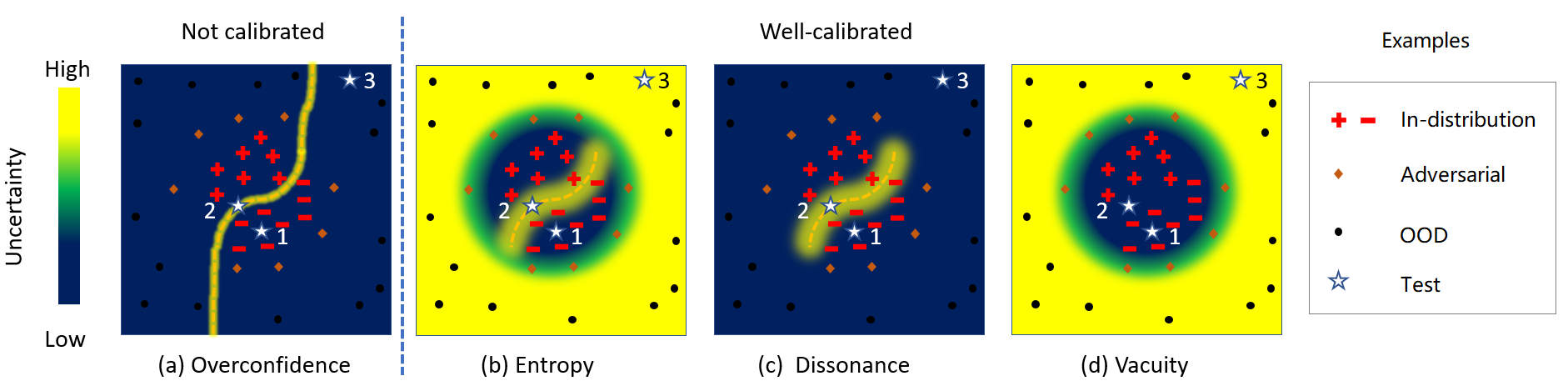}
\caption{Visualization of the predictive uncertainty in Table 1. (a) Traditional NNs with softmax function before calibration demonstrates over-confidence. (b) A well-calibrated model shows high entropy in both conflicting and OOD regions. (c) and (d) shows evidential uncertainty that decompose the uncertainty in (b) based on different root causes. The pentagrams denote the three test cases in Table 1.  }\label{fig: ideal_ood}
\end{figure*}

Besides probabilistic uncertainty and BNNs,  evidential uncertainty is proposed based on belief/evidence theory and Subjective Logic (SL)~\cite{josang2016subjective,josang2018uncertainty}.
It considers different dimensions of uncertainty, such as vacuity (i.e., lack of evidence) or dissonance (i.e., uncertainty due to conflicting evidence).
In the CV domain, \citet{sensoyuncertainty} propose evidential neural networks (ENNs) to model the uncertainty of class probabilities based on SL explicitly.  
An ENN uses the predictions as subjective opinions and learns a function that collects evidence to form the opinions by a deterministic neural network from data. 
Several works~\cite{sensoyuncertainty, zhao2019quantifying,hu2020multidimensional} improve ENNs using regularization or generative models to ensure correct uncertainty estimation towards unseen examples in image classification.
However, those methods for continuous feature space are not applicable to the discrete text.

To briefly demonstrate the motivation of our paper, we use a simple binary classification example in Table~\ref{example} and Figure~\ref{example} to answer the following questions: 
\begin{itemize}
\item Why is it necessary to calibrate predictive uncertainty?
\item What is the advantage of evidential uncertainty in OOD detection?
\item How to design a regularization method to calibrate the predictive uncertainty?
\end{itemize}

In Table~\ref{example},  we assume that a classifier is only trained on the restaurant reviews dataset and has never seen examples from other domains.  The probability denotes the prediction softmax probability.
The evidence represents historical observations, denoted by Dirichlet distributions (no evidence when $\alpha=1$).
Before calibration, the classifier predicts Sentence 3, an obvious OOD example, as positive with high confidence. 
Thus it is necessary to calibrate predictive uncertainty is to reduce over-confidence.

For a well-calibrated model, there are three common cases in predictions.  
Sentence 1 refers to correct confident classification, where we have enough evidence with no conflicts. 
Sentence 2 is vague and contains conflicting information like 'bad' and 'acceptable'. The prediction will result in equal probability because each category supports equal evidence, i.e., conflicting evidence or high dissonance. 
Finally, we lack the evidence to support our prediction for an OOD sample, Sentence 3. It results in high vacuity with Dirichlet distribution being a uniform distribution. 
The model outputs the same predictive probability for Sentence 2 and 3,  which have pretty different evidence. In this case, probabilistic uncertainty cannot distinguish the conflicting case and the out-of-distribution case. Evidential uncertainty decomposes the uncertainty base on different root causes. This explains the advantage of evidential uncertainty over probabilistic uncertainty.

Figure~\ref{fig: ideal_ood} illustrates the prediction uncertainty of neural networks in Table~\ref{example}. Assume we project the examples in a 2D space. Sentence 1 lies in the region with many negative examples. Sentence 2 lies in the boundary region. Sentence 3 is far away from the ID region. 
Figure~\ref{fig: ideal_ood} (a) represents the prediction by traditional neural networks with softmax and demonstrates over-confidence. It only assigns high uncertainty (entropy) near the classification boundary.
\citet{hein2019relu} prove that ReLU type neural networks produce arbitrary high confidence predictions far away from the training data.
Figure~\ref{fig: ideal_ood} (b) represents the predictive entropy of a well-calibrated model. 
Figure~\ref{fig: ideal_ood} (c) and (d) shows the evidential uncertainty decomposes the uncertainty in (b) based on different root causes. We observe high vacuity in OOD regions and high dissonance in ID boundary regions. Vacuity can effectively detect OOD samples from boundary ID examples because the cause of uncertainty is due to a lack of evidence.  We can distinguish sentence 3 from sentence 2 in Figure~\ref{fig: ideal_ood} (c) but not in Figure~\ref{fig: ideal_ood} (b).

Finally, in Figure~\ref{fig: ideal_ood} we also observe OOD examples and adversarial examples. Adversarial examples~\cite{szegedy2013intriguing, carlini2017towards,madry2017towards} refer to instances with small feature perturbations. A lot of studies~\cite{Jia2017adversarial, Wallace2019Triggers, Jia2019TowardsRA} use adversarial examples to evaluate and improve neural networks' robustness. 
We can use diverse outliers to calibrate the model to output high uncertainty in the OOD region~\cite{hendrycks2018deep}.  Additionally, adversarial examples can be helpful to detect OOD examples close to ID regions. 
Thus, our approach adopts a mixture of an auxiliary dataset of outliers and close adversarial examples to calibrate the predictive uncertainty.
We can easily get diverse text data as auxiliary outliers.  However,  generating adversarial examples via common gradient-based approaches is impossible in the NLP domain. 
Thus, we apply methods~\cite{stutz2019disentangling,gilmer2018adversarial,kong2020calibrated} to generate off-manifold adversarial examples from the embedding layer.

Our work provides the following {\bf key contributions} :
(\romannumeral1) We firstly apply evidential uncertainty to solve OOD detection tasks in the text classification.
(\romannumeral2) We propose an inexpensive framework that adopts both an auxiliary dataset of outliers and generated pseudo off-manifold samples to train a model with prior knowledge of a certain class, which has high vacuity for OOD samples.
(\romannumeral3) We validate our proposed method's performance via extensive experiments of OOD detection and uncertainty estimation in text classification. 
Our approach significantly outperforms all the counterparts.

\section{Preliminaries} \label{sec:background}

We briefly provide the background knowledge of evidential uncertainty and the advantage over probabilistic uncertainty.

\subsection{Subjective Opinions in SL}

A multinomial opinion in a given proposition $x$ is represented by $\omega_Y = (\bm{b}_Y, u_Y, \bm{a}_Y)$ where a domain is $\mathbb{Y} \equiv \{1, \cdots, K\}$, a random variable $Y$ takes value in $\mathbb{Y}$, $K = |\mathbb{Y}| \geq 2$ and $\omega_Y$ is given as $\sum_{y \in \mathbb{Y}} \bm{b}_Y(y) + u_Y = 1$.  $\bm{b}_Y$ denotes {\em belief mass function} over $\mathbb{Y}$. $u_Y$ denotes {\em uncertainty mass} representing {\em vacuity of evidence}. $\bm{a}_Y$ represents {\em base rate distribution} over $\mathbb{Y}$, with $\sum_y \bm{a}_Y(y)=1$.
Then the projected probability distribution of a multinomial opinion is given by:
\begin{equation} \label{eq:multinomial-projected}
\mathbf{p}_Y(y) = \bm{b}_Y(y) + \bm{a}_Y(y) u_Y,\;\;\; \forall y \in \mathbb{Y}.
\end{equation}  
Multinomial probability density over a domain of cardinality $K$ is represented by the $K$-dimensional Dirichlet PDF where the special case with $K=2$ is the Beta PDF as a binomial opinion. It denotes a domain of $K$ mutually disjoint elements in $\mathbb{Y}$ and $\alpha_Y$ the strength vector over $y \in \mathbb{Y}$ and ${\bf p}_Y$ the probability distribution over $\mathbb{Y}$.
\begin{eqnarray} \label{eq:multinomial-dir}
\mathrm{Dir}(\bm{p}_Y; {\bm \alpha}_Y) 
= \frac{1}{B({\bm \alpha}_Y)} \prod_{y \in \mathbb{Y}} \bm{p}_Y (y) ^{({\bm \alpha}_Y(y)-1)},
\end{eqnarray}
where $B({\bm \alpha}_Y)$ is a multivariate beta function as the normalizing constant, ${\bm \alpha}_Y(y) \geq 0$, and ${\bf p}_Y (y) \neq 0$ if ${\bm \alpha}_Y (y) < 1$.

We term \textit{evidence} as a measure of the amount of supporting observations collected from data in favor of a sample to be classified into a certain class. Let ${\bf r}_Y(y) \ge 0 $ be the evidence derived for the singleton $y\in \mathbb{Y}$.  The total strength ${\bm \alpha}_Y(y)$ for the  belief of each singleton $y \in \mathbb{Y}$ is given by: 
\begin{eqnarray} \label{eq:multinomial-alpha}
{\bm \alpha}_Y(y) = \bm{r}_Y(y) + \bm{a}_Y(y) W, 
\end{eqnarray}
where $W$ is a non-informative weight representing the amount of uncertain evidence and $\bm{a}_Y(y)$ is the base rate distribution.  Given the Dirichlet PDF, the expected probability distribution over $\mathbb{Y}$ is:
\begin{equation} \label{eq:multinomial-expected}
\mathbb{E}_Y(y) = \frac{{\bm \alpha}_Y (y)}{\sum_{y_i \in \mathbb{Y}} {\bm \alpha}_Y (y_i)} = \frac{\bm{r}_Y(y) + \bm{a}_Y(y) W}{W + \sum_{y_i \in \mathbb{Y}} \bm{r}_Y(y_i)},   \forall_y \in \mathbb{Y}.
\end{equation}

The observed evidence in the Dirichlet PDF can be mapped to the multinomial opinions by:
\begin{equation} \label{eq:multinomial-belief}
\bm{b}_Y(y) = \frac{\bm{r}(y)}{S}, \;
u_Y = \frac{W}{S},  
\end{equation}
where $S = \sum_{y_i \in \mathbb{Y}} {\bm \alpha}_Y(y_i)$. 
We set the base rate $\bm{a}_Y(y) = \frac{1}{K}$ and the non-informative prior weight $W = K$, and hence $\bm{a}_Y(y)\cdot  W = 1$ for each $y \in \mathbb{Y}$, as these are default values considered in subjective logic.

\subsection{Uncertainty Dimensions}

\citet{josang2018uncertainty} define multiple dimensions of a subjective opinion based on the formalism of SL. Vacuity refers to uncertainty caused by insufficient information to understand a given opinion. It corresponds to uncertainty mass, $u_Y$, of an opinion in SL as:
\begin{eqnarray}\label{eq:vacuity}
\text{Vac}({\bm \alpha}_Y) = \frac{W}{S}. 
\end{eqnarray}
Dissonance denotes when there is an insufficient amount of evidence that can clearly support a particular belief.  We observe high dissonance when the same amount of evidence is supporting multiple extremes of beliefs. 
Given a multinomial opinion with non-zero belief masses, the measure of dissonance can be obtained by:
\begin{equation}
\label{eq:belief-dissonance-multi}
\text{Diss}({\bm \alpha}_Y) = \sum\limits_{y_{i}\in \mathbb{Y}}\left(\frac{\bm{b}_{Y}(y_{i})\!\!\!\! \sum\limits_{y_j \in \mathbb{Y}\setminus y_i}\!\!\!\!\!\bm{b}_{Y}(y_{j}) \mbox{Bal}(y_{j},y_{i})}{\sum\limits_{y_j \in \mathbb{Y}\setminus y_i}\bm{b}_{Y}(y_{j})}  \right),
\end{equation} 
where the relative mass balance between a pair of belief masses $\bm{b}_{Y}(y_{j})$ and $\bm{b}_{Y}(y_{i})$ is expressed by:
\begin{equation}
\label{eq:belief-balance}
\text{Bal}(y_j,y_i)=
  \begin{cases} 
  1-\frac{|\bm{b}_{Y}(y_{j})-\bm{b}_{Y}(y_{i})|}{\bm{b}_{Y}(y_{j})+\bm{b}_{Y}(y_{i})}, 
  & \text{if $\bm{b}_{Y}(y_{j}) \bm{b}_{Y}(y_{i}) \neq 0$}\\
  0, & \text{otherwise.}
  \end{cases}
\end{equation}

The above two uncertainty measures (i.e., vacuity and dissonance) can be interpreted using class-level evidence measures of subjective opinions. 
As in Table~\ref{example}, given two classes (positive, and negative), we have three subjective opinions $\{{\bm \alpha}_1, {\bm \alpha}_2, {\bm \alpha}_3\}$, represented by the two-class evidence measures as: 
${\bm \alpha}_1 = (1, 99)$ representing low uncertainty (entropy, dissonance and vacuity) which implies high confidence in a decision making context. 
${\bm \alpha}_2 = (50, 50)$ indicating high inconclusiveness due to high conflicting evidence which gives high entropy and high dissonance,
${\bm \alpha}_3 = (1, 1)$ showing the case of high vacuity which is commonly observed in OOD samples. 
Therefore, vacuity can effectively distinguish OOD samples from boundary samples because it represents a lack of evidence.

\section{Approach} \label{sec:training-enn-gen-model}

\subsection{Calibrating Evidential Neural Networks}
ENNs \cite{sensoy2018evidential} predict the evidence vector for the predicted Dirichlet distribution instead of softmax probability. Given a sample $i$ with the input feature ${\bf x}_i \in \mathbb{R}^L$  and the ground-truth label ${\bf y}_i$,  
let $f({\bf x}_i | \Theta)$ represents the predicted evidence vector predicted by the classifier with parameters  $\Theta$. 
Then the corresponding Dirichlet distribution has
parameters ${\bm \alpha}_i = f({\bf x}_{i} | \Theta) + 1$. 
The Dirichlet density $\text{Dir}({\bf p}_i ; {\bm \alpha})$ is the prior on the Multinomial distribution $\text{Multi}({\bf y}_i | {\bf p}_i)$. Then we optimize the following sum of squared loss for classfication: 
\begin{eqnarray}
\label{eq:ennloss}
\mathcal{L}(f({\bf x}_i|\Theta), {\bf y}_i) 
=\int  \frac{\|{\bf y}_i - {\bf p}_i\|_2^2}{B({\bm \alpha}_i)} \prod_{j =1}^K p_{ij} ^{(\alpha_{ij}-1)} d{\bf p}_i \nonumber \\
=\sum_{j=1}^K (y_{ij}^2 - 2 y_{ij}\mathbb{E}[p_{ij}] +\mathbb{E}[p_{ij}^2]). 
\end{eqnarray}

Since Eq.~\eqref{eq:ennloss} only relies on class labels of training samples, it does not directly measure the quality of the predicted Dirichlet distributions. The uncertainty estimates may not be accurate.
Thus, we propose a regularization method that combines ENNs and language models to quantify evidential uncertainty in text classification tasks. 
Formally, given a set of samples 
$\mathcal{D}_\text{in} = \{({\bf x}_1, {\bf y}_1), \cdots, ({\bf x}_N, {\bf y}_N)\}$,  
where ${\bf x}_i$ refers to input embedding of sentences or documents and ${\bf y}_i$ is its label. Let $P_{out}({\bf x}, {\bf y})$ and $P_{in}({\bf x}, {\bf y})$ be the distributions of the OOD and ID samples respectively.
Let $g(\cdot)$ denote the function of the pre-trained feature extraction layers.  Let $h(\cdot)$ denote the task-specific layers. We use $\Theta$ to represent the parameters of $g$ and $h$. Then we fine-tune our model by optimizing the following loss function over the parameters $\Theta$: 
\begin{eqnarray}
\label{eq:regularize_enn}
\min_{\Theta} \mathcal{F}(\Theta) = 
\mathbb{E}_{{\bf x, \bf y}\sim P_{in}({\bf x, \bf y})}[\mathcal{L}(h \circ g({\bf x}|\Theta), {\bf y}) ] \\\nonumber
+ \, \beta_{in} \cdot \mathbb{E}_{{\bf x} \sim P_{in}({\bf x})}[\text{Vac}(h \circ g({\bf x}|\Theta))] \\\nonumber
- \, \beta_{out} \cdot \mathbb{E}_{{\bf \hat{x}} \sim P_{out}({\bf \hat{x}})}
[\text{Vac}(h \circ g({\bf \hat{x}}|\Theta))].
\end{eqnarray}

The first item refers to the vanilla classification loss of ENN Eq.~\eqref{eq:ennloss}, which ensures a reasonable estimation of the ID samples' class probabilities. The second item is to reduce the vacuity estimation on ID samples. The third item is to increase the vacuity estimation on OOD samples. $\beta_{in}$  and $\beta_{out}$ are the trade-off parameters. 
The goal of minimizing Eq.~\eqref{eq:regularize_enn} is to achieve high classification accuracy, low vacuity output for ID samples, and high vacuity output for OOD samples. 
To ensure the model's generalization to the whole data space,
the choice of effective $P_{out}$ is crucial.
Although generative models have achieved success in the CV domain \cite{sensoyuncertainty,hu2020multidimensional}, they do not apply to discrete text data.
We adopt two methods that have achieved success in the NLP domain to get effective OOD regularization:  (\romannumeral1) Using auxiliary OOD datasets;   (\romannumeral2) Generating off-manifold adversarial examples.

\subsection{Utilizing Auxiliary Datasets}

The auxiliary datasets disjointed from the test datasets can be used to calibrate the neural networks' over-confidence for unseen samples. 
A critical finding in \cite{hendrycks2018deep} is that the diversity of the auxiliary dataset is important.  \citet{hu2020multidimensional} report that the methods using diverse examples beat the methods that only use close adversarial examples~\cite{hein2019relu,sensoyuncertainty} in OOD detection in image classification.
Our empirical observations also find that randomly generated sentences (we randomly sample words and concatenate them into fake sentences) do not improve the performance. One partial explanation is that these "sentences" do not contain useful semantic information. This is similar to the CV domain, where CNN models do not extract valuable features from random pixel image samples.  
Since it is easy to get a large corpus of diverse text data, utilizing a real dataset is inexpensive and straightforward.
Let $P_{oe}({\bf \hat{x}})$ be the distribution of the OOD auxiliary dataset, the regularization can be written as:  
\begin{equation}
\label{eq:oe}
\max_{\Theta} \mathbb{E}_{{\bf \hat{x}} \sim P_{oe}({\bf \hat{x}})}
[\text{Vac}(h \circ g({\bf \hat{x}}|\Theta))] 
\end{equation}

\subsection{Utilizing Off-manifold samples }

\citet{kong2020calibrated} encourage the model to output uniform distributions on pseudo off-manifold samples to alleviate the over-confidence in OOD regions.  On the contrary, we apply off-manifold samples by enforcing the model to predict high vacuity:
\begin{equation}
\max_{\Theta} \mathbb{E}_{{\bf x^\prime} \sim P_{ad}({\bf x^\prime})}
[\text{Vac}(h \circ g({\bf x^\prime}|\Theta))]
\end{equation}
where $P_{ad}({\bf \hat{x}^\prime})$ denotes the distributions of the adversarial examples.
The off-manifold samples are generated from adding relatively large perturbations towards the outside of the data manifold.  
In our NLP tasks, the data manifold refers to the embedding space because the original text is not continuous. 
Formally, given a training ID sample (embedding) $(\mathbf{x}_i,y_i)$, we generate the off-manifold sample $\mathbf{x}_i^\prime$ by:
\begin{equation}
\label{eq:off-manifold}
\mathbf{x}^{\prime *}_i=\max _{\mathbf{x}^{\prime}_i \in \mathbb{S} \left(\mathbf{x}_i, \delta_{\text {off }}\right)} 
\mathcal{L}(h \circ g({\bf x^{\prime}_i}|\Theta), {\bf y}_i)
\end{equation}
where $\mathbb{S} \left(\mathbf{x}_i, \delta_{\text {off }}\right)$ denotes an $\ell_{\infty}$ sphere centered at $\mathbf{x}_i$ with a radius $\delta_{\text {off }}$. 
The $\delta_{\text {off }}$ is relatively large to ensure that the sphere $\delta_{\text {off }}$ lies outside of the data manifold~\cite{gilmer2018adversarial,stutz2019disentangling}. 
Then we can get pseudo off-manifold samples from $\delta_{\text {off }}$ along the adversarial direction, which is calculated from the gradient of the classification loss.

Off-manifold samples can improve the uncertainty estimation in close OOD regions. 
However, the generalization of adversarial samples relies on the diversity of the features of the training data. 
\citet{hu2020multidimensional} report that models trained on CIFAR-10 can generate better adversarial examples for regularization than models trained on SVHN~\cite{netzer2011reading}. Because CIFAR-10 contains more diverse features than SVHN, a dataset of only street numbers.
Our empirical observations find that off-manifold samples can help when combined with pre-trained transformers. However, it does not provide significant improvement in vanilla GRUs/ LSTMs.
This is consistent with the empirical study~\cite{hendrycks2020pretrained} where pre-trained transformers outperform vanilla models in generalization towards OOD regions. The embeddings of pre-trained transformers contain rich features that benefit the generated adversarial examples.
Thus following \cite{kong2020calibrated},  we evaluate off-manifold regularization on BERT~\cite{devlin2018bert}.
 
\begin{figure}[htbp]
\centering
\includegraphics[width=0.48\textwidth]{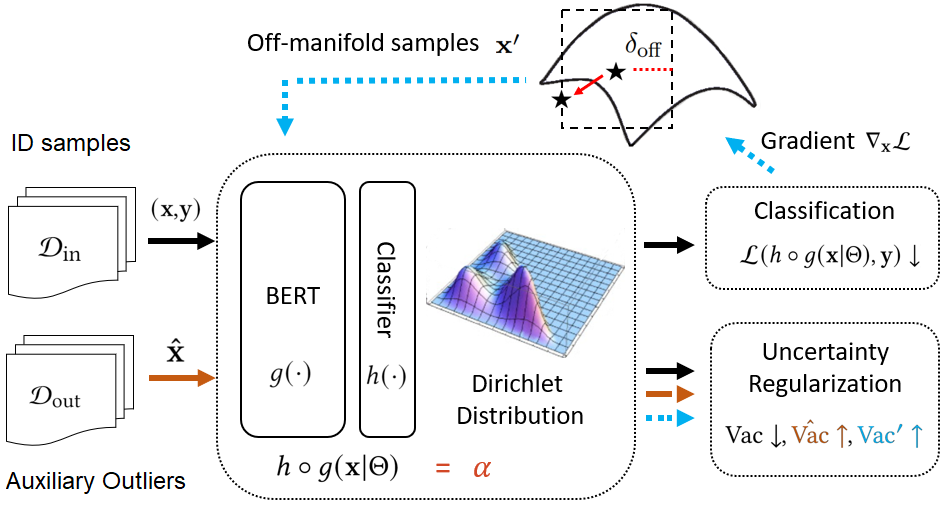}
\caption{The framework of our proposed model.}\label{fig: framework}
\end{figure}

\subsection{Mixture Regularization}

Auxiliary datasets regularization provides an overall calibration improvement, while off-manifold regularization focuses more in the close OOD region. 
We replace the last item in Eq.~\eqref{eq:regularize_enn}, which represents the uncertainty regularization for OOD data to the mixture of Eqs.~\eqref{eq:oe} and \eqref{eq:off-manifold} to get the final objective function:
\begin{eqnarray}
\min_{\Theta} \mathcal{F}(\Theta) = 
\mathbb{E}_{{\bf x, \bf y}\sim P_{in}({\bf x, \bf y})}[\mathcal{L}(h \circ g({\bf x}|\Theta), {\bf y}) ] \\\nonumber
+ \,\beta_{in} \cdot \mathbb{E}_{{\bf x} \sim P_{in}({\bf x})}[\text{Vac}(h \circ g({\bf x}|\Theta))] \\\nonumber
- \,\beta_{oe} \cdot \mathbb{E}_{{\bf \hat{x}} \sim P_{oe}({\bf \hat{x}})}
[\text{Vac}(h \circ g({\bf \hat{x}}|\Theta))] \\\nonumber
- \,\beta_{ad} \cdot \mathbb{E}_{{\bf x^\prime} \sim P_{ad}({\bf x^\prime})}
[\text{Vac}(h \circ g({\bf x^\prime}|\Theta))].
\end{eqnarray}
where $\beta_{in}$, $\beta_{oe}$, $\beta_{ad}$ denote the weight parameters of each regularization item.  
The overall framework and the detailed algorithm can be seen in Figure~\ref{fig: framework} and Algorithm~\ref{algorithm:mix}.  In each iteration, we firstly minimize the classification loss and estimated vacuity on ID samples. Then we maximize the vacuity on auxiliary outliers. Finally, we generate off-manifold samples and maximize the  vacuity estimation on them.

\section{Experiments}

We conduct OOD detection experiments on a wide range of datasets.
In each scenario, we train the model on the ID training set $\mathcal{D}_\text{in}^{\text{train}}$.
Later we evaluate the model on the ID testing set $\mathcal{D}_\text{in}^{\text{test}}$ and an OOD testing set $\mathcal{D}_\text{out}^\text{test}$ to see if the model can distinguish between ID and OOD examples. 
Our experiments consist of three parts: (\romannumeral1) We follow the work in~\cite{hendrycks2018deep} to fine-tune a simple two-layer GRU classifier~\cite{cho2014learning} using different methods. 
(\romannumeral2) Then we extend the evaluation to pre-trained language models (BERT) like~\cite{kong2020calibrated}. We report the OOD detection performance and illustrate the advantage of evidential uncertainty in (\romannumeral3) the predictive uncertainty distribution.

\begin{algorithm}[htbp]
\small
 \caption{Fine tuning our proposed mixed uncertainty model.  $f$ denotes ENN  $h \circ g (\cdot)$ with weights $\Theta$. $m$ is the batch size. $d$ is the dimension of features.}\label{algorithm:mix}
  \begin{algorithmic}[1]
    \FOR{each iteration}
        \STATE {Sample $  \{ ({\bf x}^{(i)}, y^{(i)}  )\}_{i=1}^m \!\sim\! \mathcal{D}_\text{in} $ and 
        $ \{{\bf \hat x}^{(i)}\}_{i=1}^m \!\sim\! \mathcal{D}_\text{oe} $}
        %
        %
        %
        \STATE {Update ENN by descending the gradient\\}
        \qquad  $\nabla_{\Theta}\frac{1}{m}\sum\limits_{i=1}^{m}\big{[}\mathcal{L}(f({\bf x}^{(i)}|\Theta), y^{(i)} ) 
        + \beta_{in} \text{Vac}(f({\bf x}^{(i)}|\Theta)
        \big{]}\nonumber$\\
        // \textit{Auxiliary OOD samples regularization}\\
        \STATE {Update ENN by ascending the gradient\\}
        \qquad $\nabla_{\Theta}\frac{\beta_{oe}}{m}\sum\limits_{i=1}^{N}\big[\text{Vac}(f({\bf \hat{x}}^{(i)}|\Theta)) \big]\nonumber$\\
        // \textit{Off-manifold regularization}\\
        \STATE {Initialize $\mathbf{x}_{i}^{\prime} \leftarrow \mathbf{x}_{i}+v_{i}^{\prime}$ 
        with 
        $v_{i}^{\prime} \sim 
        \mathrm{UNIF}\left[-\delta_{\text {off }}, \delta_{\text {off }}\right]^{d}$}
        \STATE {Get the gradient of the classification loss\\}
        \qquad $\Delta_{i}^{\prime} 
        \leftarrow 
        \operatorname{sign}\left(\nabla_{\mathbf{x}_{i}} \mathcal{L}(f({\bf x}_{i}|\Theta), {\bf y}_{i}))\right.$
        \STATE {Add perturbations towards off-manifold\\}
        \qquad $\mathbf{x}_{i}^{\prime} 
        \leftarrow 
        \Pi_{\left\|\mathbf{x}_{i}^{\prime}-\mathbf{x}_{i}\right\|_{\infty}=\delta_{\text {off}}}
        (\mathbf{x}_{i}^{\prime}+\delta_{\text {off}} \Delta_{i}^{\prime})$
        \STATE {Update ENN by ascending the gradient\\}
        \qquad $\nabla_{\Theta}\frac{\beta_{ad}}{m}\sum\limits_{i=1}^{N}\big[\text{Vac}(f({\bf x}^{\prime (i)}|\Theta)) \big]\nonumber$
    \ENDFOR
  \end{algorithmic}
\end{algorithm}

\subsection{Datasets}

We follow the same benchmark in~\cite{hendrycks2018deep}.
We use the same three datasets $\mathcal{D}_\text{in}$ for training and evaluating: (\romannumeral1) \textbf{20News} refers to the 20 Newsgroups dataset that contains news articles with 20 categories. 
(\romannumeral2) \textbf{SST} denotes Stanford Sentiment Treebank~\cite{socher2013recursive},  a collection of movie reviews for sentimental analysis. 
(\romannumeral3) \textbf{TREC} consists of 5, 952 individual questions with 50 classes. 
Finally, \textbf{WikiText-2} is a corpus of Wikipedia articles used for language modeling. To fairly compare with~\cite{hendrycks2018deep}, we also use its sentences as the auxiliary OOD examples $\mathcal{D}_\text{out}^{\text{train}}$ during the training.

We use the following datasets as OOD testing set $\mathcal{D}_\text{in}^{\text{test}}$: 
(\romannumeral1) \textbf{SNLI} refers to the hypotheses portion of the SNLI dataset~\cite{bowman2015large} used for natural language inference. (\romannumeral2)  \textbf{IMDB}~\cite{maas2011learning} consists of highly polar movie reviews used for sentiment classification.
(\romannumeral3) \textbf{M30K} refers to the English portion of Multi-30K~\cite{W16-3210}, a dataset of image descriptions.
(\romannumeral4) \textbf{WMT16} denotes the English portion of the test set from WMT16. 
(\romannumeral5) \textbf{Yelp} is a dataset of restaurant reviews.

\begin{table*}[htbp]
\small
\caption{The results of OOD detection using two-layer GRUs on multiple datasets. Our model (+OE) uses an auxiliary dataset for regularization. }\label{table:gru}
\centering
\begin{tabular}{cl|ccccc|ccccc|ccccc}
                          & \multicolumn{1}{c}{}                           & \multicolumn{5}{c}{FPR90 $\downarrow$}                                & \multicolumn{5}{c}{AUROC $\uparrow$}                                & \multicolumn{5}{c}{AUPR $\uparrow$}                                   \\ 
\cline{3-17}
 $\mathcal{D}_\text{in}$  & \multicolumn{1}{c|}{$\mathcal{D}_\text{out}^\text{test}$ } & MSP   & DP    & ENN   & OE             & Ours            & MSP   & DP    & ENN   & OE             & Ours            & MSP   & DP    & ENN   & OE              & Ours             \\ 
\hline
\multirow{6}{*}{\rotatebox{90}{20News}}   & SNLI                                           & 38.2  & 27.4  & 21.6  & \textbf{12.5}  & 13.2            & 87.6  & 91.4  & 92.7  & \textbf{95.1}  & 93.7            & 71.3  & 78.0    & 81.4  & \textbf{86.3}   & 71.9             \\
                          & IMDB                                           & 45    & 36.0    & 27.8  & 19.2           & \textbf{9.2}    & 79.9  & 85.1  & 88.0    & 93.6           & \textbf{96.0}     & 42.4  & 50.8  & 54.5  & 74.4            & \textbf{76.3}    \\
                          & M30K                                       & 54.5  & 42.8  & 46.0    & \textbf{3.4}   & 3.8             & 78.3  & 84.8  & 82.7  & 97.3           & \textbf{98.3}   & 46    & 60.3  & 46.3  & 93.6            & \textbf{94.9}    \\
                          & WMT16                                          & 38.7  & 29.3  & 26.7  & 1.6            & \textbf{0.8}    & 85.2  & 89.8  & 88.8  & 99.0             & \textbf{99.5}   & 57.3  & 69.2  & 56.8  & 96.6            & \textbf{98.1}    \\
                          & Yelp                                           & 45.8  & 41.2  & 39.4  & \textbf{4.0}   & 8.5             & 78.8  & 82    & 82.5  & \textbf{97.7}  & 96.5            & 37.9  & 45.3  & 41.6  & \textbf{87.8}   & 83.0               \\ 
\cline{3-17}
                          & Mean                                           & 44.44 & 35.34 & 32.3  & 8.14           & \textbf{7.1}    & 81.96 & 86.62 & 86.94 & 96.54          & \textbf{96.8}   & 50.98 & 60.72 & 56.12 & \textbf{87.74}  & 84.84            \\ 
\hline
\multirow{6}{*}{\rotatebox{90}{TREC}}     & SNLI                                           & 18.2  & 23.5  & 39.4  & 4.2            & \textbf{3.2}    & 94.0    & 89.7  & 81.7  & \textbf{98.1}  & 97.6            & 81.9  & 62.0    & 47.4  & \textbf{91.6}   & 90.0               \\
                          & IMDB                                           & 49.6  & 34.4  & 90.0    & 0.6            & \textbf{0.2}    & 78.0    & 82.4  & 45.7  & 99.3           & \textbf{99.9}   & 44.2  & 46.8  & 18.1  & 97.7   & \textbf{99.5}             \\
                          & M30K                                       & 44.2  & 33.7  & 93.6  & \textbf{0.2}   & 0.4             & 81.6  & 83.4  & 48.8  & \textbf{99.9}  & 99.6            & 44.9  & 48.1  & 19.2  & \textbf{99.3}   & 99.0               \\
                          & WMT16                                          & 50.7  & 37.9  & 93.6  & 0.6            & \textbf{0.0}    & 78.2  & 83.7  & 48.8  & 99.7           & \textbf{100}    & 42.2  & 52.4  & 19.2  & 98.9            & \textbf{99.9}    \\
                          & Yelp                                           & 50.9  & 40.1  & 83.2  & 0.2            & \textbf{0.0}    & 75.1  & 82.1  & 59.7  & 99.7           & \textbf{100}    & 37.7  & 46.8  & 24.3  & 96.3            & \textbf{100}     \\ 
\cline{3-17}
                          & Mean                                           & 42.72 & 33.92 & 79.96 & 1.16           & \textbf{0.76}   & 81.38 & 84.26 & 56.94 & 99.34          & \textbf{99.42}  & 50.18 & 51.22 & 25.64 & 96.76           & \textbf{97.68}   \\ 
\hline
\multirow{6}{*}{\rotatebox{90}{SST}}      
& SNLI   & 57.3  & 48.5  & 42.4  & 33.4           & \textbf{21.1}   & 75.7  & 76.8  & 86.0    & 86.8           & \textbf{91.4}   & 36.2  & 35.0    & 47.0    & 52.0              & \textbf{61.7}    \\
                          
& IMDB                                           & 83.0    & 85.8  & 93.6  & 32.6           & \textbf{25.5}   & 54.4  & 56.2  & 43.7  & 85.8           & \textbf{91.8}   & 19.0    & 21.3  & 15.7  & 51.3            & \textbf{76.8}    \\
& M30K                                       & 79.6  & 82.1  & 99.6  & \textbf{31.6}  & 34.3            & 59.5  & 58.1  & 32.5  & 88.3           & \textbf{89.2}   & 21.7  & 21.1  & 14.7  & 58.7            & \textbf{80.2}    \\
& WMT16                                          & 68.8  & 67.9  & 97.5  & 21.2           & \textbf{7.2}    & 66.5  & 69.1  & 50.6  & 91.7           & \textbf{96.8}   & 25.9  & 28.9  & 24.5  & 66.5            & \textbf{93.6}    \\
& Yelp                                           & 82.4  & 85.9  & 96.4  & \textbf{10.9}  & 13.6            & 53.1  & 55.1  & 35.3  & 93.4           & \textbf{95.9}   & 18.0    & 19.8  & 14.1  & 61.4            & \textbf{88.8}    \\ 
\cline{3-17}
                          & Mean & 74.22 & 74.04 & 85.9  & 25.94 & \textbf{20.34}  & 61.84 & 63.06 & 49.62 & 89.2       
                          & \textbf{93.02}  & 24.16 & 25.22 & 23.2  & 57.98           & \textbf{80.22}   \\
\hline
\end{tabular}
\end{table*}

\subsection{Comparing Schemes}

We compare several recent methods for qualifying uncertainty or OOD detection in text categorization. 
(\romannumeral1) \textbf{MSP} refers to maximum softmax probability, a baseline work of OOD detection~\cite{hendrycks2016baseline}.
(\romannumeral2) \textbf{DP} refers to Monte Carlo Dropout~\cite{gal2016dropout}, which applies dropout at train and test time. We run ten it times and use the average MSP as the uncertainty score.
(\romannumeral3) \textbf{TS} is a post-hoc calibration method by temperature scaling~\cite{guo2017calibration}. We fine-tune the temperature parameter via the validation set.
(\romannumeral4) \textbf{MRC} denotes Manifold Regularization Calibration~\cite{guo2017calibration}, which adopts on- and off-manifold regularization to improve the calibration of BERT. 
(\romannumeral5)  \textbf{OE} refers to Outlier Exposure~\cite{hendrycks2018deep} that enforces uniform confidence on an auxiliary OOD dataset.
(\romannumeral6)  \textbf{ENN}~\cite{sensoy2018evidential} is our base classifier, which uses deep learning models to explicitly model SL uncertainty. 
Most of the baselines with softmax function use the negative of maximum softmax scores ($-max_c fc(x)$) as the uncertainty score, which is similar to predictive entropy.  ENN uses predictive entropy. Our proposed model uses vacuity as the detection score. 


\subsection{Metrics}

We consider the following metrics in~\cite{hendrycks2016baseline,hendrycks2018deep}: The area under the receiver operating characteristic curve (\textbf{AUROC}),  the area under the precision-recall curve (\textbf{AUPR}) and the False Alarm Rate at 90\% Recall (\textbf{FAR90}).
Higher AUROC indicates a higher probability that a positive example has a higher score than a false example, which means better accuracy.
AUPR is similar to AUROC, but it also considers the positive class's base rate. Higher AUPR is better. FAR90 measures the probability that a false example raises a false alarm, assuming that 90\% of all positive examples are detected.  Lower FAR90 is better.

For the GRU experiments, we use the source code of MSP and OE  in~\cite{hendrycks2018deep}.  We follow the same pre-processing steps and the base rate of $\mathcal{D}_\text{out}^\text{test}$ to $\mathcal{D}_\text{in}^\text{test}$ is 1:5 in each scenario.
We implement ENN, DP, and our model based on the same two-layer GRUs. We pre-train the base classifier for five epochs and fine-tune five more epochs for OE and our model using WikiText-2.  Except for DP, we pre-train it for ten epochs to ensure the same accuracy as others.  We evaluate our model with auxiliary datasets regularization  (+OE). 

For the experiments on BERT, we follow the same setting in \cite{kong2020calibrated}, which also contains the implementation of multiple baselines.  We still set the base rate of $\mathcal{D}_\text{out}^\text{test}$  to 1:5 to be consistent with the previous experiments. We construct sequence classifiers with one linear layer on top of the pooled output of a pre-trained uncased BERT base model. Then we fine-tune it with different models for ten epochs. We evaluate auxiliary datasets regularization (+OE), adversarial regularization (+AD), and the mixture method (MIX).

We fairly train all the baselines with their default parameters and report the average results. In the GRU experiments, we set $\beta_{in} = 0.1$, $\beta_{oe} = 1$, $\mathrm{batch\_size}=128$, $\mathrm{learning\_rate}=1e^{-4}$ in Adam optimizer of our model in all the experiments, which were fine-tuned considering the performance of both the OOD detection and ID classification accuracy. 
For the experiments on BERT, we set  $\beta_{oe} = 1$ in all +OE and MIX,  $\beta_{ad} = 1$ in all +AD, $\mathrm{learning\_rate}=5e^{-5}$ in Adam optimizer in all experiments.  But we use slightly different $\beta_{in}$ for each $\mathcal{D}_\text{in}^{\text{train}}$, which is fine-tuned considering the accuracy and vacuity from the validation ID set.
For more details, refer to Section 4.7 and our source code~\footnote{\url{https://github.com/snowood1/BERT-ENN}}.

 \begin{table*}[htbp]
\small
\setlength{\tabcolsep}{2.5pt}
\centering
\caption{The results of OOD detection using BERT on multiple datasets. Our model (MIX) applies both an auxiliary dataset and off-manifold adversarial samples for regularization. } \label{table:bert}
\begin{tabular}{ll|cccccc|cccccc|cccccc}
                                             & \multicolumn{1}{l}{}                           & \multicolumn{6}{c}{FPR90 $\downarrow$ }                          & \multicolumn{6}{c}{AUROC $\uparrow$ }                             & \multicolumn{6}{c}{AUPR $\uparrow$ }                               \\ 
\cline{3-20}
\multicolumn{1}{c}{$\mathcal{D}_\text{in}$ } & \multicolumn{1}{c|}{$\mathcal{D}_\text{out}^\text{test}$ } & MSP   & DP    & TS    & MRC   & OE             & Ours             & MSP   & DP    & TS    & MRC   & OE              & Ours             & MSP   & DP    & TS    & MRC   & OE              & Ours              \\ 
\hline
\multirow{6}{*}{\rotatebox{90}{20News}}            & SNLI                                           & 16.6  & 22.1  & 14.5  & 0.8   & \textbf{0.0}  & \textbf{0.0}   & 94.4  & 92.7  & 95.2  & 99.3  & \textbf{100.0 } & \textbf{100.0 } & 85.1  & 80.0  & 87.8  & 97.6  & \textbf{100.0 }           & \textbf{100.0 }  \\
                                             & IMDB                                           & 16.3  & 19.0  & 14.9  & 15.4  & 6.3            & \textbf{0.0}   & 92.4  & 91.0  & 93.5  & 94.5  & 97.8            & \textbf{99.7 }  & 70.6  & 65.0  & 76.6  & 81.8  & 93.5            & \textbf{99.6 }   \\
                                             & M30K                                       & 16.7  & 21.1  & 14.9  & 2.5   & \textbf{0.0}  & \textbf{0.0}   & 94.0  & 91.7  & 94.9  & 99.0  & \textbf{100.0 } & \textbf{100.0 } & 82.9  & 75.8  & 86.4  & 96.5  & \textbf{100.0 } & \textbf{100.0 }  \\
                                             & WMT16                                          & 21.1  & 23.6  & 19.4  & 10.9  & \textbf{0.0}  & \textbf{0.0}   & 91.3  & 90.4  & 92.2  & 97.0  & \textbf{100.0 } & \textbf{100.0 } & 73.9  & 71.2  & 77.8  & 90.4  & \textbf{100.0 } & 99.9             \\
                                             & Yelp                                           & 26.9  & 29.5  & 26.0  & 23.4  & 14.3           & \textbf{0.0}   & 86.7  & 84.5  & 87.6  & 89.0  & 95.3            & \textbf{98.7 }  & 50.6  & 43.2  & 53.9  & 58.8  & 86.0            & \textbf{98.2 }   \\ 
\cline{3-20}
                                             & Mean                                           & 19.52 & 23.05 & 17.93 & 10.60 & 4.13           & \textbf{0.00 }  & 91.75 & 90.10 & 92.68 & 95.74 & 98.62           & \textbf{99.69 } & 72.61 & 67.05 & 76.51 & 85.01 & 95.90           & \textbf{99.53 }  \\ 
\hline
\multirow{6}{*}{\rotatebox{90}{TREC}}              & SNLI                                           & 89.8  & 89.8  & 90.0  & 79.6  & 6.2            & \textbf{0.0}   & 42.7  & 45.5  & 42.6  & 62.6  & 95.6            & 99.3            & 18.0  & 18.5  & 18.2  & 27.4  & 93.9            & \textbf{99.4 }   \\
                                             & IMDB                                           & 43.6  & 45.0  & 44.6  & 37.0  & \textbf{0.0}  & \textbf{0.0}   & 74.6  & 73.9  & 75.0  & 83.4  & 99.3            & \textbf{99.7 }  & 31.3  & 30.5  & 32.6  & 54.0  & 98.7            & \textbf{99.5 }   \\
                                             & M30K                                       & 89.8  & 90.0  & 90.4  & 88.2  & 89.2           & \textbf{0.0}   & 32.3  & 34.6  & 32.9  & 53.9  & 84.8            & \textbf{100.0 } & 14.6  & 15.0  & 14.8  & 21.1  & 83.8            & \textbf{100.0 }  \\
                                             & WMT16                                          & 35.4  & 29.6  & 30.0  & 23.8  & \textbf{0.0}  & \textbf{0.0}   & 84.0  & 84.5  & 84.5  & 92.7  & \textbf{99.3 }  & \textbf{99.3 }  & 45.9  & 45.7  & 48.5  & 78.0  & 98.5            & \textbf{98.8 }   \\
                                             & Yelp                                           & 29.0  & 28.4  & 29.8  & 20.6  & \textbf{0.0}  & \textbf{0.0}   & 83.7  & 83.9  & 83.8  & 91.4  & 97.7            & \textbf{98.9 }  & 45.8  & 45.0  & 46.8  & 73.0  & 96.6            & \textbf{98.6 }   \\ 
\cline{3-20}
                                             & Mean                                           & 57.52 & 56.56 & 56.96 & 49.84 & 19.08          & \textbf{0.00 }  & 63.46 & 64.50 & 63.78 & 76.79 & 95.34           & \textbf{99.44 } & 31.14 & 30.95 & 32.19 & 50.69 & 94.31           & \textbf{99.27 }  \\ 
\hline
\multirow{6}{*}{\rotatebox{90}{SST}}               & SNLI                                           & 57.6  & 58.4  & 57.6  & 48.1  & 31.5           & \textbf{22.1 }  & 75.3  & 73.2  & 75.3  & 75.7  & 90.2            & \textbf{93.4 }  & 35.8  & 32.0  & 35.8  & 31.9  & 67.9            & \textbf{78.7 }   \\
                                             & IMDB                                           & 67.0  & 63.0  & 67.0  & 15.8  & 49.9           & \textbf{0.4 }   & 70.8  & 69.4  & 70.8  & 93.9  & 83.5            & \textbf{97.7 }  & 30.8  & 28.0  & 30.8  & 75.4  & 61.0            & \textbf{96.1 }   \\
                                             & M30K                                       & 42.4  & 45.9  & 42.4  & 41.6  & 26.6           & \textbf{20.3 }  & 80.8  & 78.8  & 80.8  & 79.2  & 91.5            & \textbf{94.2 }  & 41.5  & 38.1  & 41.5  & 35.6  & 70.2            & \textbf{79.1 }   \\
                                             & WMT16                                          & 56.6  & 57.6  & 56.6  & 58.3  & \textbf{52.1 } & 70.4            & 79.2  & 77.5  & 79.2  & 74.2  & \textbf{81.2 }  & 77.2            & 41.3  & 37.9  & 41.3  & 31.2  & 55.1            & \textbf{56.0 }   \\
                                             & Yelp                                           & 62.3  & 60.8  & 62.3  & 44.4  & 39.3           & \textbf{3.5 }   & 71.9  & 70.1  & 71.9  & 86.0  & 86.9            & \textbf{97.0 }  & 30.3  & 28.5  & 30.3  & 59.0  & 60.9            & \textbf{94.4 }   \\ 
\cline{3-20}
                                             & Mean                                           & 57.18 & 57.14 & 57.18 & 41.66 & 39.89          & \textbf{23.34 } & 75.59 & 73.79 & 75.59 & 81.80 & 86.65           & \textbf{91.92 } & 35.92 & 32.90 & 35.92 & 46.62 & 63.01           & \textbf{80.88 }  \\
\hline
\end{tabular}
\end{table*}

\begin{table*}[htbp]
\centering
\small
\caption{ The ablation study of different regularization's effects on BERT-ENNs. We show vanilla ENNs, with auxiliary outliers (+OE), with off-manifold examples (+AD), and with the mixture of both methods (MIX). We also list the best counterpart OE.}
\label{table:ablation}
\begin{tabular}{ll|ccccc|ccccc|ccccc}
                                             & \multicolumn{1}{l}{}                                       & \multicolumn{5}{c}{ FPR90 $\downarrow$ }                                   & \multicolumn{5}{c}{ AUROC $\uparrow$ }                                      & \multicolumn{5}{c}{ AUPR $\uparrow$ }                                        \\ 
\cline{3-17}
\multicolumn{1}{c}{$\mathcal{D}_\text{in}$ } & \multicolumn{1}{c|}{$\mathcal{D}_\text{out}^\text{test}$ } & OE           & ENN   & +OE             & +AD            & MIX              & OE             & ENN   & +OE             & +AD            & MIX             & OE             & ENN   & +OE             & +AD            & MIX              \\ 
\hline
\multirow{6}{*}{\rotatebox{90}{20News}}                     & SNLI                                                       & \textbf{0.0} & 61.2  & \textbf{0.0}    & 6.0            & \textbf{0.0}     & \textbf{100.0} & 80.6  & \textbf{100.0}  & 96.8           & \textbf{100.0}  & \textbf{100.0} & 64.2  & \textbf{100.0}  & 87.6           & \textbf{100.0}   \\
                                             & IMDB                                                       & 6.3          & 94.6  & 0.7             & 7.8            & \textbf{0.0}     & 97.8           & 53.3  & 98.2            & 94.6           & \textbf{99.7}   & 93.5           & 32.9  & 96.9            & 90.2           & \textbf{99.6}    \\
                                             & M30K                                                       & \textbf{0.0} & 59.1  & \textbf{0.0}    & 5.3            & \textbf{0.0}     & \textbf{100.0} & 79.3  & \textbf{100.0}  & 96.7           & \textbf{100.0}  & \textbf{100.0} & 58.2  & \textbf{100.0}  & 85.3           & \textbf{100.0}   \\
                                             & WMT16                                                      & \textbf{0.0} & 85.9  & \textbf{0.0}    & 11.5           & \textbf{0.0}     & \textbf{100.0} & 68.4  & \textbf{100.0}  & 93.6           & \textbf{100.0}  & \textbf{100.0} & 49.1  & \textbf{100.0}  & 84.6           & 99.9             \\
                                             & Yelp                                                       & 14.3         & 74.7  & 0.6             & 10.3           & \textbf{0.0}     & 95.3           & 62.6  & 97.3            & 94.7           & \textbf{98.7}   & 86.0           & 25.0  & 96.1            & 81.8           & \textbf{98.2 }   \\ 
\cline{3-17}
                                             & Mean                                                       & 4.13         & 75.10 & 0.25            & 8.20           & \textbf{0.00}    & 98.62          & 68.85 & 99.10           & 95.30          & \textbf{99.69}  & 95.90          & 45.87 & 98.59           & 85.89          & \textbf{99.53}   \\ 
\hline
\multirow{6}{*}{\rotatebox{90}{TREC}}                        & SNLI                                                       & 6.2          & 42.6  & \textbf{0.0}    & 67.4           & \textbf{0.0}     & 95.6           & 86.0  & \textbf{100.0}  & 68.6           & 99.3            & 93.9           & 75.3  & \textbf{100.0}  & 42.0           & 99.4             \\
                                             & IMDB                                                       & \textbf{0.0} & 74.0  & \textbf{0.0}    & \textbf{0.0}   & \textbf{0.0}     & 99.3           & 53.5  & \textbf{100.0}  & 99.3           & 99.7            & 98.7           & 21.2  & \textbf{100.0}  & 98.2           & 99.5             \\
                                             & M30K                                                       & 89.2         & 36.4  & \textbf{0.0}    & 67.2           & \textbf{0.0}     & 84.8           & 91.0  & 98.6            & 59.2           & \textbf{100.0}  & 83.8           & 81.6  & 98.8            & 27.5           & \textbf{100.0}   \\
                                             & WMT16                                                      & \textbf{0.0} & 81.0  & \textbf{0.0}    & 29.8           & \textbf{0.0}     & 99.3           & 47.5  & \textbf{99.6}   & 91.3           & 99.3            & 98.5           & 19.5  & \textbf{99.1}   & 78.4           & 98.8             \\
                                             & Yelp                                                       & \textbf{0.0} & 70.0  & \textbf{0.0}    & 19.4           & \textbf{0.0}     & 97.7           & 63.7  & \textbf{99.4}   & 94.9           & 98.9            & 96.6           & 27.2  & \textbf{99.4}   & 92.2           & 98.6             \\ 
\cline{3-17}
                                             & Mean                                                       & 19.08        & 60.80 & \textbf{0.00 }  & 36.76          & \textbf{0.00}    & 95.34          & 68.34 & \textbf{99.52}  & 82.66          & 99.44           & 94.31          & 44.98 & \textbf{99.47}  & 67.66          & 99.27            \\ 
\hline
\multirow{6}{*}{\rotatebox{90}{SST}}                       & SNLI                                                       & 31.5         & 64.6  & \textbf{14.6}   & 38.3           & 22.1             & 90.2           & 74.7  & \textbf{95.2}   & 85.9           & 93.4            & 67.9           & 37.0  & \textbf{82.4}   & 59.3           & 78.7             \\
                                             & IMDB                                                       & 49.9         & 68.0  & 76.5            & 13.3           & \textbf{0.4}     & 83.5           & 63.1  & 79.5            & 95.9           & \textbf{97.7}   & 61.0           & 23.8  & 66.5            & 91.8           & \textbf{96.1}    \\
                                             & M30K                                                       & 26.6         & 55.5  & \textbf{7.4}    & 25.7           & 20.3             & 91.5           & 84.3  & \textbf{95.9}   & 90.7           & 94.2            & 70.2           & 46.9  & \textbf{81.8}   & 69.6           & 79.1             \\
                                             & WMT16                                                      & 52.1         & 79.8  & 62.6            & \textbf{51.1}  & 70.4             & 81.2           & 59.1  & 77.5            & \textbf{82.1}  & 77.2            & 55.1           & 24.5  & 52.4            & \textbf{56.8}  & 56.0             \\
                                             & Yelp                                                       & 39.3         & 68.3  & 29.6            & 26.1           & \textbf{3.5}     & 86.9           & 63.8  & 90.7            & 92.7           & \textbf{97.0}   & 60.9           & 24.9  & 72.7            & 85.6           & \textbf{94.4}    \\ 
\cline{3-17}
                                             & Mean                                                       & 39.89        & 67.23 & 38.13           & 30.92          & \textbf{23.34 }  & 86.65          & 69.00 & 87.76           & 89.47          & \textbf{91.92}  & 63.01          & 31.41 & 71.16           & 72.61          & \textbf{80.88}   \\
\hline
\end{tabular}
\end{table*}

\subsection{Out-of-Distribution Detection} 
 
In Table~\ref{table:gru}, our model on GRU significantly outperforms other approaches on SST and achieves the overall best results on TREC. Except on 20News, OE slightly outperforms ours.
One partial explanation is that simple GRUs can not handle accuracy and uncertainty estimation simultaneously when handling longer texts.  
The average accuracy of all the models is only 73\%, which indicates that the models have not learned the correct evidence. 

Table~\ref{table:bert} shows that pre-trained models still suffer from over-confidence.  DP does not outperform MSP, which is consistent with \cite{Vernekar2019out} that  MC Dropout only measures uncertainty in ID settings. 
TS still replies on softmax probability and tune its temperature parameter on the validation (ID) set. Thus TS does not generalize well in unseen data. 
Therefore, effective OOD detection models require regularization from OOD examples.  
OE using a diverse real auxiliary dataset beats MRC that adopts adversarial examples, except in the close OOD setting SST vs. IMDB. 
Our model (MIX) applies both regularizations and beats both of them. 

Table~\ref{table:ablation} further analyzes the contribution of each regularization. Both +OE and +AD improve the performance of vanilla ENN.  +OE outperforms the baseline OE. This indicates the effectiveness of evidential uncertainty when using the same regularization. 
While +OE provides an overall improvement, 
+AD is especially effective in distinguishing close OOD examples. For example, in SST vs. IMDB and SST vs. Yelp, both cases involve movies or reviews.  
In sum, applying the mixture of both regularizations achieves the overall stable best performance.

\begin{figure*}[htpb]
\centering

\subfloat[20News: ID vs. OOD 
\label{fig:20news_ID_OOD}]{
  \includegraphics[width=0.65\columnwidth,keepaspectratio]{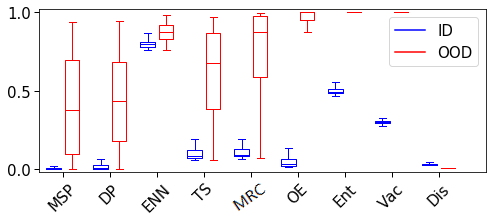}
}
\quad
\subfloat[TREC: ID vs. OOD 
\label{fig:trec_ID_OOD}]{
  \includegraphics[width=0.65\columnwidth,keepaspectratio]{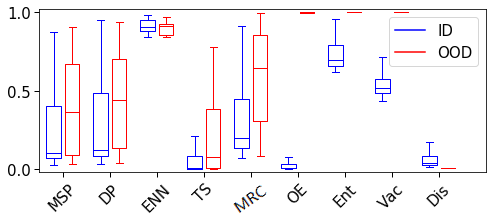}
}
\quad
\subfloat[SST: ID vs. OOD 
\label{fig:sst_ID_OOD}]{
  \includegraphics[width=0.65\columnwidth,keepaspectratio]{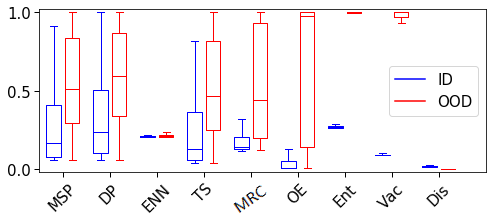}
}
\\
\subfloat[20News: Successful vs. Failed predictions 
\label{fig:20news_succ_fail}]{
  \includegraphics[width=0.65\columnwidth,keepaspectratio]{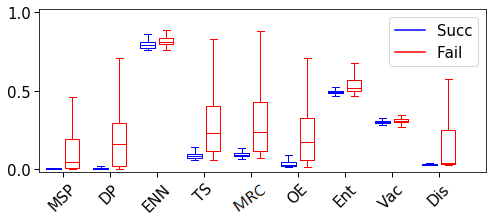}
}
\quad
\subfloat[TREC: Successful vs. Failed predictions 
\label{fig:trec_succ_fail}]{
  \includegraphics[width=0.65\columnwidth,keepaspectratio]{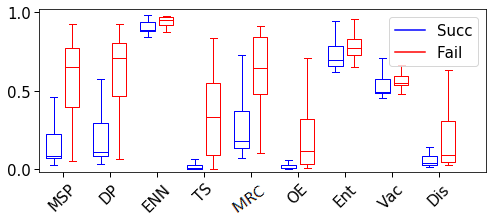}
}
\quad
\subfloat[SST: Successful vs. Failed predictions 
\label{fig:sst_succ_fail}]{
  \includegraphics[width=0.65\columnwidth,keepaspectratio]{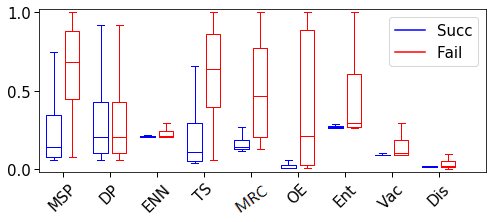}
}

\caption{Top row: The boxplots of predictive uncertainty of different models on different  $\mathcal{D}_\text{in}^\text{test}$ vs. examples from all the four OOD datasets $\mathcal{D}_\text{out}^\text{test}$. 
Bottom row: The boxplots of predictive uncertainty of successful and failed predictions in different $\mathcal{D}_\text{in}^\text{test}$.
Our model uses entropy (Ent), vacuity (Vac), dissonance (Dis) as a  measure of uncertainty, while other models use entropy. }\label{fig:boxplot}

\end{figure*}

\begin{figure}[htbp]
\centering
\includegraphics[width=0.5\columnwidth]{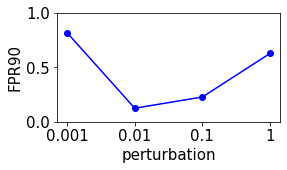}
\caption{The OOD detection performance of our model (+AD) using off-manifold adversarial regularization with different  $\delta_{\text{off }}$ in the scenario SST ($\mathcal{D}_\text{in}$) vs. IMDB ($\mathcal{D}_\text{out}^\text{test}$). }\label{fig: eps out}
\end{figure}

\subsection{Predictive Uncertainty Distribution} 

We use boxplots to show the uncertainty distribution of different models deployed on BERT in Fig.\ref{fig:boxplot}.  Baselines use entropy as a measure of uncertainty.  Our proposed model use vacuity (\textbf{Vac}) and the square root of dissonance (\textbf{Dis}) ranged from [0, 1].  We also show the output of our entropy (\textbf{Ent}). 
The top row shows the predictive uncertainty in $\mathcal{D}_\text{in}^\text{test}$ and compares them to those for all the OOD datasets. We concatenate all the five OOD datasets as OOD examples in these experiments.
The bottom row shows different models’ predictive uncertainty for correct and mis-classified examples in  $\mathcal{D}_\text{in}^\text{test}$.
OE is the best counterpart in OOD detection. However, OE fails to give a distinct separation between ID and OOD data on SST.
Besides, all the counterparts predict high uncertainty for misclassified ID samples the same as OOD samples. Thus they will misclassify some of the boundary ID samples as OOD samples. 
On the contrary, our model decomposes the uncertainty into vacuity and dissonance.  High vacuity is observed only in the OOD region.
The boundary ID samples will have higher dissonance but low vacuity. 
This explains the advantage of adopting vacuity in distinguish between boundary ID and OOD examples.

\subsection{Parameter Study}

The most important parameters are $\delta_\text{off}=0.01$ and $\beta_\text{oe}=1$.
$\delta_\text{off}$ influences the performance of adversarial regularization greatly. 
We find that  $\delta_\text{off}=0.01$ achieves the best performance across all of our experiments.  
Figure \ref{fig: eps out} shows the FPR90 of our model using off-manifold regularization (+AD) in the scenario SST ($\mathcal{D}_\text{in}$) vs. IMDB ($\mathcal{D}_\text{out}^\text{test}$). 
We observe the same performance in all the other scenarios. When $\delta_\text{off}$ is too small, the generated samples might be too close to the manifold and may harm the confidence of the ID region.  
Too much perturbation leads to ineffective samples for regularization. 

We also compare the effect of the weights of different regularization terms in the mixture formula.
We find that +OE provides an overall improvement in calibration, and we simply set $\beta_{oe} = 1$.
We try different $\beta_{ad} = 1$ or 0.1 to better distinguish close OOD examples. 
$\beta_{in}$ is tuned via the validation ID set within three possible values 0, 0.01 and 0.1.
Since the first item in Eq.~\eqref{eq:regularize_enn} already assigns considerable confidence in training samples during the classification process, it also reduces ID samples' vacuity.  Large $\beta_{in}$ may also affect the accuracy.  Therefore we only use a small $\beta_{in}$ to scale the vacuity of ID examples slightly. The summary of different weights can be seen in Table~\ref{table:parameters}.

\begin{table}[htpb]
\small
\centering
\caption{Hyper-parameters for BERT-ENNs}
\label{table:parameters}
\begin{tabular}{cc|ccccc}
\toprule
                        &     &  $\beta_{in}$    & $\beta_{oe}$  & $\beta_{ad}$      \\ 
\midrule
\multirow{3}{*}{20News} & +OE & 0.1  & 1   & -     \\
                        & +AD & 0    & -   & 1     \\
                        & MIX & 0    & 1   & 0.1    \\
\midrule
\multirow{3}{*}{TREC}   & +OE & 0    & 1   & -      \\
                        & +AD & 0    & -   & 1      \\
                        & MIX & 0    & 1   & 0.1    \\
\midrule
\multirow{3}{*}{SST}    & +OE & 0.01 & 1   & -      \\
                        & +AD & 0.01 & -   & 1     \\
                        & MIX & 0.01 & 1   & 1      \\
\bottomrule

\end{tabular}
\end{table}

\section{Related work}

Our study is related to uncertainty qualification~\cite{blundell2015weight,gal2016dropout,sensoy2018evidential}, OOD detection~\cite{hendrycks2016baseline,hendrycks2018deep} and confidence calibration~\cite{guo2017calibration,Thulasidasan2019mixup,kong2020calibrated}. We have discussed the NLP applications of these fields in the Introduction.

Other baselines not included in our experiments include Deep Ensemble~\cite{lakshminarayanan2017simple}, 
which average the softmax outputs of five models with different initialization.  A recent empirical study \cite{ovadia2019canshift} proves that Deep Ensemble performs better than Dropout and Temperature Scaling under dataset shift of NLP tasks using LSTM~\cite{schmidhuber1997long}. However, fine-tuning multiple pre-trained transformer models is computationally expensive.  Besides, the advantage of our considered baseline OE over this method has been reported in \cite{meinke2019towards}.  Therefore we do not consider this method as a baseline in our paper.
Another line of work, Stochastic Variational Bayesian Inference \cite{blundell2015weight,louizos2017multiplicative,wen2018flipout} can be applied to CNN models but hard to be applied in other architectures such as LSTMs~\cite{ovadia2019canshift}. \citet{sensoy2018evidential, hu2020multidimensional} also prove the advantage of ENNs over multiple Stochastic Variational Bayesian Inference methods.

\section{Conclusion}
Qualifying uncertainty is essential for reliable classification, but less work has been studied in the NLP domain.
We firstly apply evidential uncertainty based on SL to solve OOD detection in the text classification. We combine ENNs and language models to measure vacuity and dissonance.
Our proposed model uses auxiliary datasets of outliers and off-manifold samples to train a model with prior knowledge of a certain class, which has high vacuity for OOD samples.
Extensive experiments show that our approach significantly outperforms all the counterparts.

\begin{acks}
\small{
The research reported herein was supported in part by NSF awards DMS-1737978, DGE-2039542, OAC-1828467, OAC-1931541, and DGE-1906630, ONR awards N00014-17-1-2995 and N00014-20-1-2738, Army Research Office Contract No. W911NF2110032 and IBM faculty award (Research).}
\end{acks}

\bibliographystyle{ACM-Reference-Format}
\bibliography{sample-base}


\end{document}